# Block matching algorithm for motion estimation based on Artificial Bee Colony (ABC)

**Erik Cuevas[1,a], Daniel Zaldívar[a], Marco Pérez-Cisneros[a], Humberto Sossa[b] and Valentín Osuna[b]**

[a]Departamento de Electrónica
Universidad de Guadalajara, CUCEI
Av. Revolución 1500, C.P 44430, Guadalajara, Jal, México
{ erik.cuevas, daniel.zaldivar, marco.perez}@cucei.udg.mx
[b]Centro de Investigación en Computación-IPN
Av. Juan de Dios Batiz S/N, Mexico, D. F. MEXICO
hsossa@cic.ipn.mx

**Abstract**

Block matching (BM) motion estimation plays a very important role in video coding. In a BM approach, image frames in a video sequence are divided into blocks. For each block in the current frame, the best matching block is identified inside a region of the previous frame, aiming to minimize the sum of absolute differences (SAD). Unfortunately, the SAD evaluation is computationally expensive and represents the most consuming operation in the BM process. Therefore, BM motion estimation can be approached as an optimization problem, where the goal is to find the best matching block within a search space. The simplest available BM method is the full search algorithm (FSA) which finds the most accurate motion vector through an exhaustive computation of SAD values for all elements of the search window. Recently, several fast BM algorithms have been proposed to reduce the number of SAD operations by calculating only a fixed subset of search locations at the price of poor accuracy. In this paper, a new algorithm based on Artificial Bee Colony (ABC) optimization is proposed to reduce the number of search locations in the BM process. In our algorithm, the computation of search locations is drastically reduced by considering a fitness calculation strategy which indicates when it is feasible to calculate or only estimate new search locations. Since the proposed algorithm does not consider any fixed search pattern or any other movement assumption as most of other BM approaches do, a high probability for finding the true minimum (accurate motion vector) is expected. Conducted simulations show that the proposed method achieves the best balance over other fast BM algorithms, in terms of both estimation accuracy and computational cost.

## 1. Introduction

Video coding is currently employed for a considerable number of applications including fixed and mobile telephony, real-time video conferencing, DVD and high-definition digital television. Motion Estimation (ME) is an important part of any video coding system, since it can achieve significant compression by exploiting the temporal redundancy that commonly exists in a video sequence. Several ME methods have been studied seeking for a complexity reduction at video coding such as block matching (BM) algorithms, parametric-based models [1], optical flow[2] and pel-recursive techniques [3]. Among such methods, BM seems to be the most popular technique due to its effectiveness and simplicity for both software and hardware implementations [4]. In order to reduce the computational complexity in ME, many BM algorithms have been proposed and employed at implementations for several video compression standards such as MPEG-4 [5] and H.264 [6].

In BM algorithms, the video frames are partitioned into non-overlapping blocks of pixels. Each block is predicted from a block of equal size in the previous frame. In particular, for each block at the current frame, the algorithm aims for the best matching block within a search window from the previous frame, while minimizing a certain matching metric. The most used matching measure is the sum of absolute differences (SAD) which is computationally expensive and represents the most consuming operation in the BM process.

---
[1] Corresponding author, Tel +52 33 1378 5900, Ext. 27714, E-mail: erik.cuevas@cucei.udg.mx

The best matching block thus represents the predicted block, whose displacement from the previous block is represented by a transitional motion vector (MV). Therefore, BM is essentially an optimization problem whose goal is to find the best matching block within a search space.

The full search algorithm (FSA) [7] is the simplest block-matching algorithm that can deliver the optimal estimation solution with respect to a minimal matching error as it checks all candidates one at a time. However, such exhaustive search and the full-matching error calculation at each checking point, yields an extremely computational expensive BM method that seriously constraints its use for real-time video applications.

In order to decrease the computational complexity of the BM process, several BM algorithms have been proposed considering the following three techniques: (1) using a fixed pattern: the search operation is conducted over a fixed subset of the total search window. The Three Step Search (TSS) [8], the New Three Step Search (NTSS) [9], the Simple and Efficient TSS (SES) [10], the Four Step Search (4SS) [11] and the Diamond Search (DS) [12], all represent some of its well-known examples. Although such approaches have been algorithmically considered as the fastest, they are not able to eventually match the dynamic motion-content, sometimes delivering false motion vectors (image distortions). (2) Reducing the search points: the algorithm chooses as search points only those locations that iteratively minimize the error-function (SAD values). This category includes the Adaptive Rood Pattern Search (ARPS) [13], the Fast Block Matching Using Prediction (FBMAUPR) [14], the Block-based Gradient Descent Search (BBGD) [15] and the Neighbourhood Elimination algorithm (NE) [16]. Such approaches assume that the error-function behaves monotonically, holding well for slow-moving sequences but failing for other kind of movements in video sequences [17], making the algorithm prone to get trapped into local minima. (3) Decreasing the computational overhead for every search point: the matching cost (SAD operation) is replaced by a partial or a simplified version that features less complexity. The New pixel-Decimation (ND) [18], the Efficient Block Matching Using Multilevel Intra, the Inter-Sub-blocks [9] and the Successive Elimination Algorithm [19], all assume that all pixels within each block, move by the same finite distance and a good estimate of the motion can be obtained through only a fraction of the pixel pool. However, since only a fraction of pixels enters into the matching computation, the use of such regular sub-sampling techniques can seriously affect the accuracy of the detection of motion vectors due to noise or illumination changes.

Another popular group of BM algorithms employ spatio-temporal correlation by using neighboring blocks in the spatial and temporal domain in order to predict MVs. The main advantage of such algorithms is that they alleviate the local minimum problem to some extent as the new initial or predicted search center is usually closer to the global minimum and therefore the chance of getting trapped in a local minimum decreases. This idea has been incorporated by many fast-block motion estimation algorithms such as the Enhanced Predictive Zonal Search (EPZS) [20] and the UMHexagonS [21]. However, the information delivered by the neighboring blocks occasionally conduces to false initial search points producing distorted motion vectors. Such problem is typically caused by the movement of very small objects contained in the image sequences [22].

Alternatively, evolutionary approaches such as genetic algorithms (GA) [23] and particle swarm optimization (PSO) [24] are well known for delivering the location of the global optimum in complex optimization problems. Despite of such fact, only few evolutionary approaches have specifically addressed the problem of BM, such as the Light-weight Genetic Block Matching (LWG) [25], the Genetic Four-step Search (GFSS) [26] and the PSO-BM [27]. Although these methods support an accurate identification of the motion vector, their spending times are very long in comparison to other BM techniques.

Karaboga has recently presented one bee-swarm algorithm for solving numerical optimization problems which is known as the Artificial Bee Colony (ABC) method [28]. Inspired by the intelligent foraging behaviour of honeybee swarm, the ABC algorithm consists of three essential components: food source positions, nectar amount and different honey bee classes. Each food source position represents a feasible solution for the problem under consideration and the nectar amount of a food source represents the quality of such solution corresponding to its fitness value. Each class of bee symbolizes one particular operation for generating new candidate food source positions (candidate solutions). The ABC algorithm starts by producing a randomly distributed initial population (food source locations). After initialization, an objective function evaluates whether such candidates represent an acceptable solution (nectar amount) or not. Guided by the values of such objective function, the candidate solutions are evolved through different ABC operations (honey bee types). When the fitness function (nectar amount) cannot be further improved after a maximum number of cycles is reached, its related food source is assumed to be abandoned and replaced by a new randomly chosen food source location. The performance of ABC algorithm has been compared to other optimization methods such as Genetic Algorithms (GA), Differential Evolution (DE) and Particle Swarm Optimization (PSO) [29,30]. The results showed that ABC can produce optimal solutions and thus is more effective than other methods in several optimization problems. Such characteristics have motivated the use of ABC to solve different sorts of engineering problems such as signal processing [31,32], flow shop scheduling [33], structural inverse analysis [34], clustering [35,36], vehicle path planning [37] and image processing [38,39].

One particular difficulty in applying ABC to real-world applications is about its demand for a large number of fitness evaluations before delivering a satisfying result. However, fitness evaluations are not always straightforward in many real-world applications as either an explicit fitness function does not exist or the fitness evaluation is computationally very expensive. Furthermore, since random numbers are involved in the calculation of new individuals, they may encounter same positions (repetition) that have been visited by other individuals at previous iterations, particularly when individuals are confined to a small area.

The problem of considering expensive fitness evaluations has already been faced in the field of evolutionary algorithms (EA) and is better known as fitness approximation [40]. In such approach, the idea is to estimate the fitness value of so many individuals as it is possible instead of evaluating the complete set. Such estimations are based on an approximate model of the fitness landscape. Thus, the individuals to be evaluated and those to be estimated are determined following some fixed criteria which depend on the specific properties of the approximate model [41]. The models involved at the estimation can be built during the actual EA run, since EA repeatedly samples the search space at different points [42]. There are many possible approximation models and several have already been used in combination with EA (e.g. polynomials [43], the kriging model [44], the feed-forward neural networks that includes multi-layer Perceptrons [45] and radial basis-function networks [46]). These models can be either global, which make use of all available data or local which make use of only a small set of data around the point where the function is to be approximated. Local models, however, have a number of advantages [42]: they are well-known and suitably established techniques with relatively fast speeds. Moreover, they employ the most remarkable information for the estimation of newer points: the closest neighbors.

In this paper, a new algorithm based on ABC is proposed to reduce the number of search locations in the BM process. The algorithm uses a simple fitness calculation approach which is based on the Nearest Neighbor Interpolation (NNI) algorithm in order to estimate the fitness value (SAD operation) for several candidate solutions (search locations). As a result, the approach can substantially reduce the number of function evaluations yet preserving the good search capabilities of ABC. The proposed method achieves the best balance over other fast BM algorithms, in terms of both estimation accuracy and computational cost.

The overall paper is organized as follows: Section 2 holds a brief description about the ABC algorithm. In Section 3, the fitness calculation strategy for solving the expensive optimization problem is presented. Section 4 provides backgrounds about the BM motion estimation issue while Section 5 exposes the final BM algorithm as a combination of ABC and the fitness calculation strategy. Section 6 demonstrates experimental results for the proposed approach over standard test sequences and some conclusions are drawn in Section 7.

## 2. Artificial Bee Colony (ABC) algorithm

The ABC algorithm assumes the existence of a set of operations that may resemble some features from the honey bee behavior. For instance, each solution within the search space includes a parameter set representing food source locations. The "fitness value" refers to the food source quality which is strongly linked to the food's location. The process mimics the bee's search for valuable food sources yielding an analogous process for finding the optimal solution.

*2.1 Biological bee profile*

The minimal model for a honey bee colony consists of three classes: employed bees, onlooker bees and scout bees. The employed bees will be responsible for investigating about food sources and sharing the information to recruit onlooker bees, which in turn, will make a decision on choosing food sources considering such information. The food source holding a higher quality will have a larger chance to be selected by onlooker bees than one showing a lower quality. An employed bee, whose food source is rejected as low quality by employed and onlooker bees, will change to a scout bee to randomly search for new food sources. Therefore, the exploitation is driven by employed and onlooker bees while the exploration is maintained by scout bees. The implementation details of such bee-like operations in the ABC algorithm are described in the next sub-section.

*2.2 Description of the ABC algorithm*

Resembling other swarm based approaches, the ABC algorithm is an iterative process. It starts with a population of randomly generated solutions or food sources. The following three operations are applied until a termination criterion is met [30]:

1. Send the employed bees.
2. Select the food sources by the onlooker bees.
3. Determine the scout bees.

*2.2.1 Initializing the population*

The algorithm begins by initializing $N_p$ food sources. Each food source is a *D*-dimensional vector containing the parameter values to be optimized which are randomly and uniformly distributed between the pre-specified lower initial parameter bound $x_j^{low}$ and the upper initial parameter bound $x_j^{high}$.

$$x_{j,i} = x_j^{low} + \text{rand}(0,1) \cdot (x_j^{high} - x_j^{low});$$
$$j = 1, 2, \ldots, D; \quad i = 1, 2, \ldots, N_p. \tag{1}$$

with *j* and *i* being the parameter and individual indexes respectively. Hence, $x_{j,i}$ is the *j*th parameter of the *i*th individual.

*2.2.2 Send employed bees*

The number of employed bees is equal to the number of food sources. At this stage, each employed bee generates a new food source in the neighborhood of its present position as follows:

$$v_{j,i} = x_{j,i} + \phi_{j,i}(x_{j,i} - x_{j,k});$$
$$k \in \{1, 2, \ldots, N_p\}; j \in \{1, 2, \ldots, D\} \tag{2}$$

$x_{j,i}$ is a randomly chosen *j* parameter of the *i*th individual and *k* is one of the $N_p$ food sources, satisfying the conditional $i \neq k$. If a given parameter of the candidate solution $v_i$ exceeds its predetermined boundaries, that parameter should be adjusted in order to fit the appropriate range. The scale factor $\phi_{j,i}$ is a random number between $[-1,1]$. Once a new solution is generated, a fitness value representing the profitability associated to a particular solution is calculated. The fitness value for a minimization problem can be assigned to each solution $v_i$ by the following expression

$$fit_i = \begin{cases} \dfrac{1}{1+J_i} & \text{if } J_i \geq 0 \\ 1 + abs(J_i) & \text{if } J_i < 0 \end{cases} \tag{3}$$

where $J_i$ is the objective function to be minimized. A greedy selection process is thus applied between $v_i$ and $x_i$. If the nectar amount (fitness) of $v_i$ is better, then the solution $x_i$ is replaced by $v_i$, otherwise $x_i$ remains.

*2.2.3 Select the food sources by the onlooker bees*

Each onlooker bee selects one of the proposed food sources depending on their fitness value which has been recently defined by employed bees. The probability that a food source will be selected can be obtained from the equation below:

$$Prob_i = \frac{fit_i}{\sum_{i=1}^{N_p} fit_i} \tag{4}$$

where $fit_i$ is the fitness value of the food source $i$, which is related to the objective function value ($J_i$) corresponding to the food source $i$. The probability of a food source being selected by the onlooker bees increases as the fitness value of a food source increases. After the food source is selected, onlooker bees will go to the selected food source and select a new candidate food source position inside the neighborhood of the selected food source. The new candidate food source can be expressed and calculated by Eq. (2). In case that the nectar amount i.e. fitness of the new solution, is better than before, such position is held, otherwise the old solution remains.

*2.2.4 Determine the scout bees*

If a food source $i$ (candidate solution) cannot be further improved through a predetermined number "*limit*" of trials, the food source is assumed to be abandoned, and the corresponding employed or onlooker bee becomes a scout. A scout bee explores the searching space with no previous information i.e. the new solution is generated randomly as in Eq.(1). In order to verify if a candidate solution has reached the predetermined "*limit*", a counter $A_i$ is assigned to each food source $i$. Such counter is incremented as a consequence of a bee-operation failing to improve the food source's fitness.

## 3. Fitness approximation method

Evolutionary algorithms that use fitness approximation aim to find the global minimum of a given function by considering only a small number of function evaluations and a large number of estimations. Such algorithms commonly employ alternative models of the function landscape in order to approximate the actual fitness function. The application of this method requires that the objective function fulfils two conditions: a heavy computational overhead and a small number of dimensions (up to five) [47].

Recently, several fitness estimators have been reported in the literature [43-46] in which the number of function evaluations is considerably reduced to hundreds, dozens or even less. However, most of these methods produce complex algorithms whose performance is conditioned to the quality of the training phase and the learning algorithm in the construction of the approximation model.

In this paper, we explore the use of a local approximation scheme based on the nearest-neighbor-interpolation (NNI) for reducing the function evaluation number. The model estimates fitness values based on previously evaluated neighboring individuals which have been stored during the evolution process. At each generation, some individuals of the population are evaluated through the accurate (actual) fitness function while other remaining individuals are only estimated. The positions to be accurately evaluated are determined either by their proximity to the best individual or regarding their uncertain fitness value.

### 3.1 Updating the individual database

In a fitness approximation method, every evaluation or estimation of an individual produces one data point (individual position and fitness value) that is potentially considered for building the approximation model during the evolution process. In our proposed approach, all seen-so-far evaluations are kept in a history array **T** which is employed to select the closest neighbor and to estimate the fitness value of a newer individual. Since all data are preserved and potentially available for their use, the model construction is faster because only the most relevant data points are actually used by the approach.

### 3.2 Fitness calculation strategy

This section discusses details about the strategy to decide which individuals are to be evaluated or estimated. The proposed fitness calculation scheme estimates most of fitness values to reduce the computational overhead at each generation. In the model, those individuals lying nearer to the best fitness value holder, currently registered in the array **T** (rule 1), are evaluated by using the actual fitness function. Such individuals are relevant as they possess a stronger influence on the evolution process than others. On the other hand, evaluation is also compulsory for those individuals lying in a region of the search space which has been unexplored so far (rule 2). The fitness values for such individuals are uncertain since there is no close reference (close points contained in **T**) to calculate their estimates. The rest of the individuals, lying in a region of the search space that contains enough previously calculated points, must be estimated using the NNI (rule 3). This rule indicates that the fitness value for such individuals must be estimated by assigning the fitness value from the nearest individual stored in **T**.

Therefore, the fitness calculation model follows three important rules to evaluate or estimate fitness values:

1. ***Exploitation rule (evaluation).*** If a new individual (search position) $P$ is located closer than a distance $d$ with respect to the nearest individual $L_q$ ($q = 1, 2, 3, \ldots, m$; where $m$ is the number of elements contained in **T**) with a fitness value $F_{L_q}$ that corresponds to the best fitness value seen-so-far, then the fitness value of $P$ is evaluated using the actual fitness function. Figure 1b draws this rule procedure.

2. ***Exploration rule (evaluation).*** If a new individual $P$ is located further away than a distance $d$ with respect to the nearest individual $L_q$, then its fitness value is evaluated by using the actual fitness function. Figure 1c outlines the rule procedure.

3. ***NNI rule (estimation).*** If a new individual $P$ is located closer than a distance $d$ with respect to the nearest individual $L_q$, whose fitness value $F_{L_q}$ does not correspond to the best fitness value, then its fitness value is estimated by assigning the same fitness that $L_q$ ($F_P = F_{L_q}$). Figure 1d sketches the rule procedure.

The $d$ value controls the trade-off between the evaluation and the estimation of search locations. Typical values of $d$ range from 2 to 4. Fig. 1 illustrates the procedure of fitness computation for a new solution (point $P$). In the problem (Fig. 1a), it is considered the fitness function $f$ with respect to two parameters ($x_1, x_2$), where the individuals database array **T** contains five different elements ($L_1, L_2, L_3, L_4, L_5$) and their corresponding fitness values ($F_{L_1}, F_{L_2}, F_{L_3}, F_{L_4}, F_{L_5}$). Figures 1(b) and 1(c) show the fitness evaluation ($f(x_1, x_2)$) of the new solution $P$, following the rule 1 and rule 2 respectively, whereas Fig. 1(d) presents the fitness estimation of $P$ using the NNI approach which has been laid by rule 3.

The proposed fitness calculation strategy, seen from an optimization perspective, favors the exploitation and exploration in the search process. For the exploration, the method evaluates the fitness function of new search locations which have been located far away from previously calculated positions. Additionally, it also estimates those which are closer. For the exploitation, the proposed method evaluates the actual fitness function of those new individuals which are located nearby the position that holds the minimum fitness value seen-so-far, aiming to improve its minimum. After several simulations, the value of *d*=3 has shown the best balance between the exploration and exploitation inside the search space (in the context of a BM application); thus it has been used in this paper.

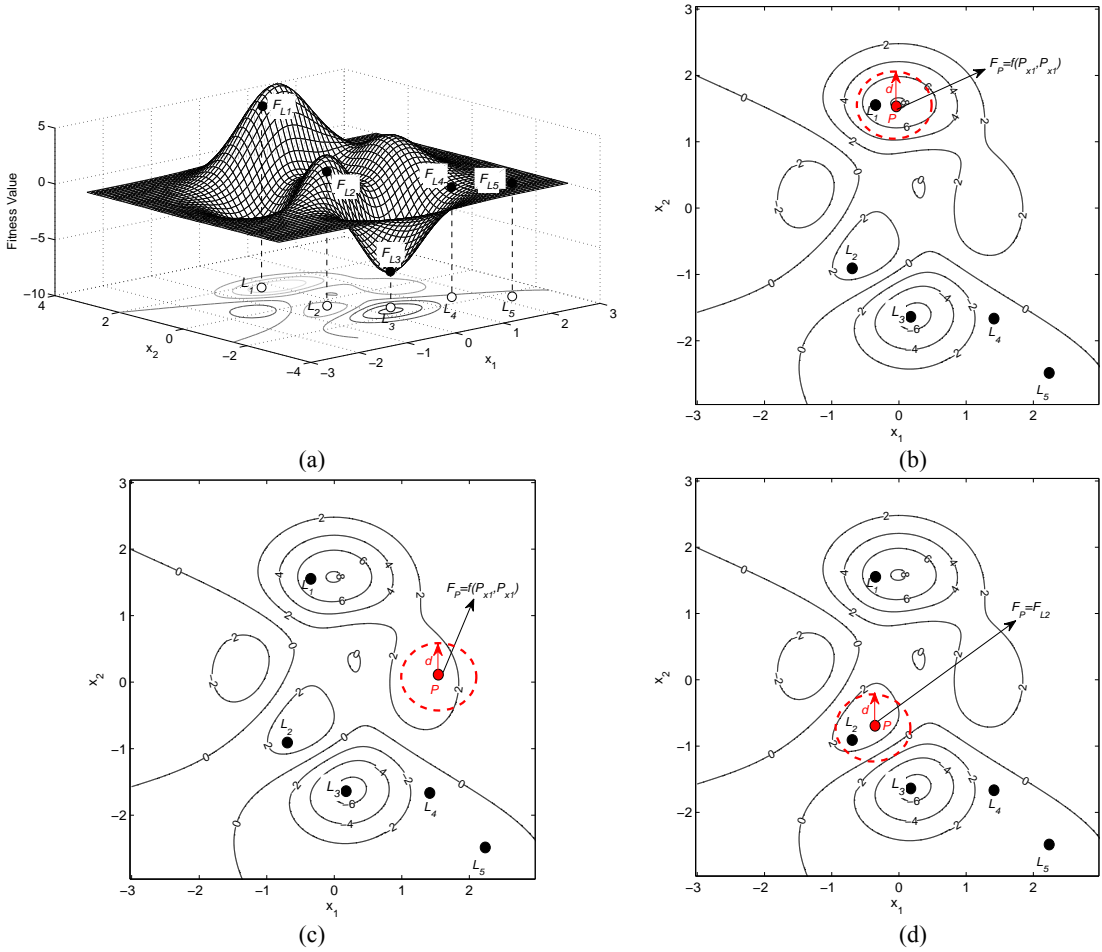

**Fig. 1.** The fitness calculation strategy. (a) Fitness function and the history array **T** content. (b) According to the rule 1, the individual (search position) $P$ is evaluated as it is located closer than a distance $d$ with respect to the best individual $L_1$.
(c) According to the rule 2, the search point $P$ is evaluated because there is no reference within its neighborhood. (d) According to rule 3, the fitness value of $P$ is estimated by the NNI-estimator, assigning $F_P = F_{L_2}$

*3.3 Proposed ABC optimization method*

In this section, the incorporation of the fitness calculation strategy to the ABC algorithm is presented. Only the fitness calculation scheme shows the difference between the conventional ABC and the enhanced approach. In the modified ABC, only some individuals are actually evaluated (rules 1 and 2) at each generation. The fitness values for the rest are estimated using the NNI-approach (rule 3). The estimation is executed using individuals that have been already stored in the array **T**.

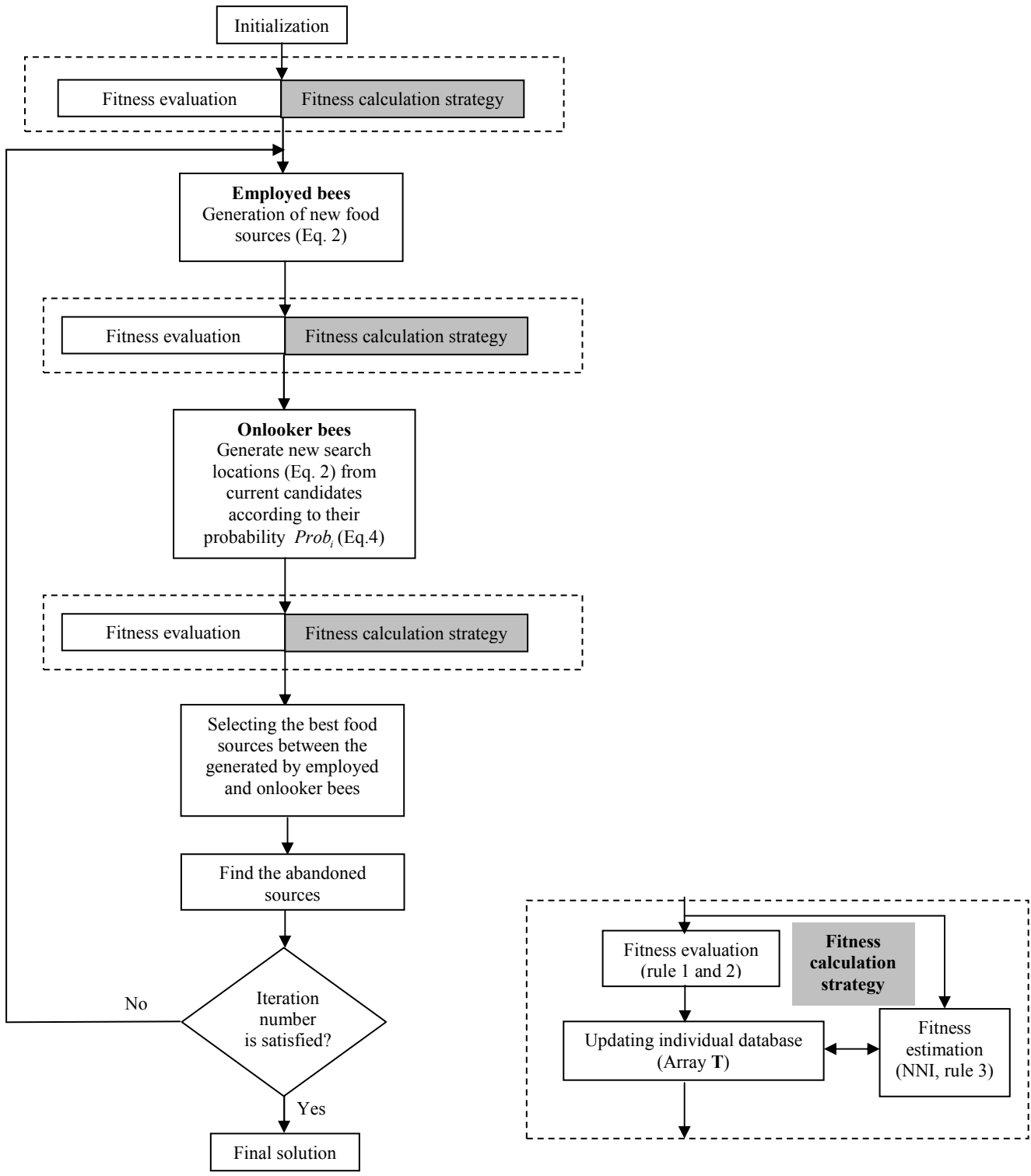

(a)                  (b)

**Fig. 2.** Proposed ABC optimization method. (a) Differences between the conventional ABC and the modified ABC. (b) The proposed fitness calculation strategy

Fig. 2 shows the difference between the conventional ABC and its modified version. In Fig 2a, it is clear that the way in which the fitness value is calculated represents the only difference between both methods. In the original ABC, each individual is evaluated according to traditional evolutionary algorithms by using the objective function. On the other hand, the modified ABC, the proposed fitness calculation strategy for obtaining the fitness value has been employed. Fig. 2b shows the components of the fitness calculation strategy: the fitness evaluation, the fitness estimation and the updating of the individual database. As a result, the ABC approach can substantially reduce the number of function evaluations yet preserving the good search capabilities of ABC.

## 4. Motion estimation and block matching

For motion estimation through a BM algorithm, the current frame of an image sequence $I_t$ is divided into non-overlapping blocks of $N$x$N$ pixels. For each template block in the current frame, the best matched block within a search window ($S$) of size $(2W+1)$x$(2W+1)$ in the previous frame $I_{t-1}$ is determined, where $W$ is the maximum allowed displacement. The position difference between a template block in the current frame and the best matched block in the previous frame is called the Motion Vector (MV) (see Fig. 3). Under such perspective, BM can be approached as an optimization problem aiming for finding the best MV within a search space.

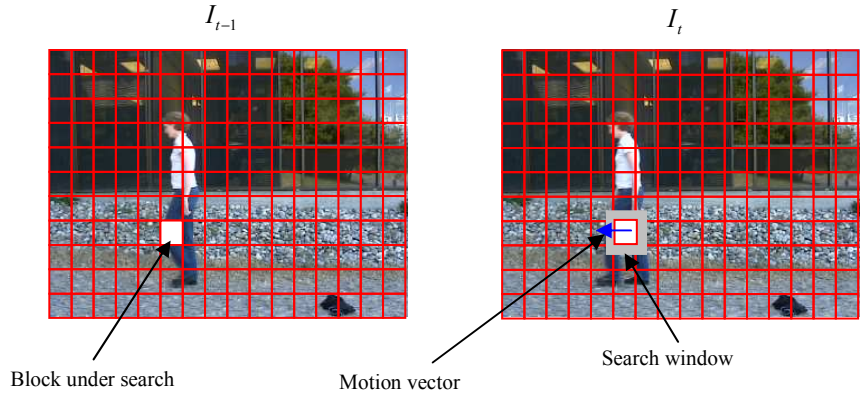

Fig. 3. Block matching procedure.

The most well-known criterion for BM algorithms is the sum of absolute differences (SAD). It is defined in Eq.(5) considering a template block at position $(x, y)$ in the current frame and the candidate block at position $(x+\hat{u}, y+\hat{v})$ in the previous frame $I_{t-1}$:

$$\text{SAD}(\hat{u},\hat{v}) = \sum_{j=0}^{N-1}\sum_{i=0}^{N-1} |g_t(x+i, y+j) - g_{t-1}(x+\hat{u}+i, y+\hat{v}+j)| \quad (5)$$

where $g_t(\cdot)$ is the gray value of a pixel in the current frame $I_t$ and $g_{t-1}(\cdot)$ is the gray level of a pixel in the previous frame $I_{t-1}$. Therefore, the MV in $(u,v)$ is defined as follows:

$$(u,v) = \arg \min_{(u,v)\in S} \text{SAD}(\hat{u},\hat{v}) \quad (6)$$

where $S = \{(\hat{u},\hat{v}) | -W \leq \hat{u},\hat{v} \leq W$ and $(x+\hat{u}, y+\hat{v})$ is a valid pixel position $I_{t-1}\}$.

In the context of BM algorithms, the FSA is the most robust and accurate method to find the MV. It tests all possible candidate blocks from $I_{t-1}$ within the search area to find the block with the minimum SAD. For the maximum displacement of $W$, the FSA requires $(2W+1)^2$ search points. For instance, if the maximum displacement $W$ is $\mp 7$, the total search points are 225. Each SAD calculation requires $2N^2$ additions and the total number of additions for the FSA to match a 16×16 block is 130,560. Such computational requirement makes the application of FSA difficult for real time tasks.

## 5. BM algorithm based on ABC with the estimation strategy

FSA finds the global minimum (the accurate MV), considering all locations within the search space *S*. Nevertheless, the approach has a high computational cost for practical use. In order to overcome such a problem, many fast algorithms have been developed yielding only a poorer precision than the FSA. A better BM algorithm should spend less computational time on the search and get more accurate motion vectors (MVs).

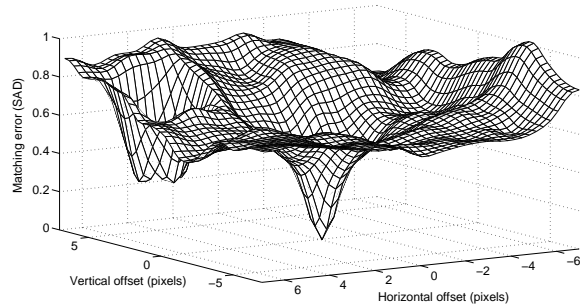

**Fig. 4.** Common non-uni-modal error surface with multiple local minimum error points

The BM algorithm proposed in this paper is comparable to the fastest algorithm yet delivering a similar precision to the FSA approach. Since most of fast algorithms use a regular search pattern or assume a characteristic error function (uni-modal) for searching about the motion vector, they may get trapped into local minima considering that, for many cases (i.e. complex motion sequences), an uni-modal error is no longer valid. Fig. 4 shows a typical error surface which has been computed around the search window for a fast-moving sequence. On the other hand, the proposed BM algorithm uses a non-uniform search pattern for locating global minimum distortion. Under the effect of the ABC operators, the search locations vary from generation to generation, avoiding to get trapped into a local minimum. Besides, since the proposed algorithm uses a fitness calculation strategy for reducing the evaluation of the SAD values, it requires fewer search positions.

In the algorithm, the search space *S* consists of a set of 2-D motion vectors $\hat{u}$ and $\hat{v}$ representing the *x* and *y* components of the motion vector, respectively. The particle is thus defined as:

$$P_i = \{\hat{u}_i, \hat{v}_i \mid -W \leq \hat{u}_i, \hat{v}_i \leq W\} \qquad (7)$$

where each particle *i* represents a possible motion vector. In this paper, all search windows that are considered for the simulations are set to ±8 and ±16 pixels. Both configurations have been selected in order to compare their results to other approaches presented in the literature.

*5.1 Initial population*

The first step in ABC optimization is to generate an initial group of individuals. The standard literature of evolutionary algorithms generally suggests the use of random solutions as initial population, considering the absence of knowledge about the problem [48]. However, Li [49] and Xiao [50] demonstrated that the use of solutions generated through some domain knowledge to set the initial population (i.e., non-random solutions) can significantly improve its performance. In order to obtain appropriate initial solutions (based on knowledge), an analysis over the motion vector distribution has been conducted. After considering several sequences (see Table 1 and Fig. 8), it can be seen that 98% of the MVs are found to lie at the origin of the search window for a slow-moving sequence such as the one at *Container*, whereas complex motion sequences, such as the *Carphone* and the *Foreman* examples, have only 53.5% and 46.7% of their MVs in the central search region. The *Stefan* sequence, showing the most complex motion content, has only 36.9%. Figure 6 shows the surface of the MV distribution for the *Foreman* and the *Stefan*. On the other hand, although it is less evident, the MV distribution of several sequences shows small peaks at some locations lying away from the center as they are contained inside a rectangle which is shown in Fig. 5(b) and 5(d) by a white overlay. Real-world moving sequences concentrate most of the MVs under a limit due to the motion continuity principle [16]. Therefore, in this paper, initial solutions are selected from five fixed locations which represent points showing the higher concentration in the MV distribution, just as it is shown by Figure 6.

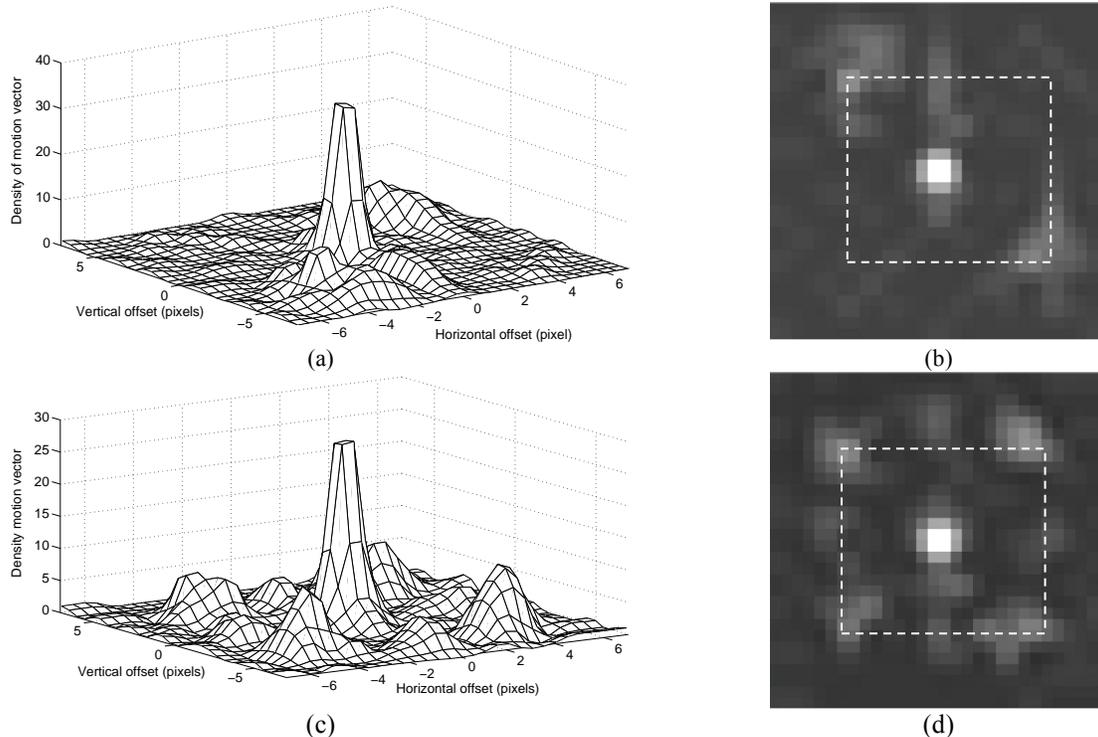

**Fig. 5.** Motion vector distribution for *Foreman* and Stefan sequences. (a)-(b) MV distribution for the *Foreman* sequence. (c)-(d) MV distribution for the *Stefan* sequence.

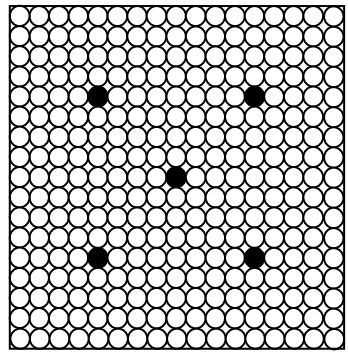

**Fig. 6.** Fixed pattern of five elements in the search window of ±8 which are to be used as initial solutions

Since the largest portion of the BM translations between neighboring video frames tend to behave as it is shown by the pattern of Fig. 6, such initial solutions are used in order to accelerate the ABC method. This consideration is taken regardless of the employed search window size (±8 or ±16).

*5.2 The ABC-BM algorithm*

The goal of our BM-approach is to reduce the number of evaluations of the SAD values (actual fitness function) avoiding any performance loss and achieving an acceptable solution. The ABC-BM method is listed below:

**Step 1:** Set the ABC parameters (*limit*=10).

**Step 2:** Initialize the population of 5 individuals $\mathbf{P} = \{P_1, \ldots, P_5\}$ using the pattern that has been shown in Fig. 6 and the individual database array $\mathbf{T}$, as an empty array. Clear all counters $A_i$ ($i \in 1, \ldots, 5$) too.

**Step 3:** Compute the fitness values for each individual according to the fitness calculation strategy presented in Section 3. Since all individuals of the initial population fulfil rule 2 conditions, they are evaluated through the actual fitness function by calculating the actual SAD values.

**Step 4:** Update new evaluations in the individual database array **T.**

**Step 5:** Modify each element of the current population **P** (search locations) as stated by Eq. 2, in order to produce a new solution vector $\mathbf{E} = \{E_1, \ldots, E_5\}$. Likewise, update all counters $A_i$.

**Step 6:** Compute fitness values for each individual by using the fitness calculation strategy presented in Section 3.

**Step 7:** Update new evaluations in the individual database array **T.**

**Step 8:** Calculate the probability value $Prob_i$ for each candidate solution that is to be used as a preference index by onlooker bees.

**Step 9:** Generate new search locations (using the Eq. 2) from **E** according to their probability $Prob_i$ (send onlooker bees to their selected food source), in order to produce the solution vector $\mathbf{O} = \{O_1, \ldots, O_5\}$. Likewise, update counters $A_i$.

**Step 10:** Compute fitness values for each individual by using the fitness calculation strategy presented in Section 3.

**Step 11:** Update new evaluations in the individual database array **T.**

**Step 12:** Generate the new population **P**. In case that the fitness value (evaluated or approximated) of the new solution $O_i$, is better than the solution $E_i$, such position is selected as an element of **P**, otherwise the solution $E_i$ is chosen.

**Step 13:** If the solution counter $A_i$ exceeds the number "*limit*" of trials, the solution *i* is re-started and generated randomly using Eq. 1. Likewise, the counter $A_i$ is cleared.

**Step 14:** Determine the best individual of the current new population. If the new fitness (SAD) value is better than the old one, then update $\hat{u}_{best}$, $\hat{v}_{best}$.

**Step 15:** If the number of target iterations has been reached (four in the case of a search window size of ±8 and eight for ±16), then the MV is $\hat{u}_{best}$, $\hat{v}_{best}$; otherwise go back to Step 5.

The proposed ABC-BM algorithm considers different search locations during the complete optimization process with 45 in the case of a search window of ±8 and 85 for ±16, with 4 and 8 different iterations respectively. However, only a few of search locations are to be evaluated using the actual fitness function (between 5 and 14 in the case of a search window of ±8 and between 7 and 22 for the case of ±16) while all other remaining positions are just estimated. Figure 7 shows two search-pattern examples that have been generated by the ABC-BM approach. Such patterns exhibit the evaluated search-locations (rule 1 and 2) in white-cells, whereas the minimum location is marked in black. Grey-cells represent those that have been estimated (rule 3) or not visited at all, during the optimization process.

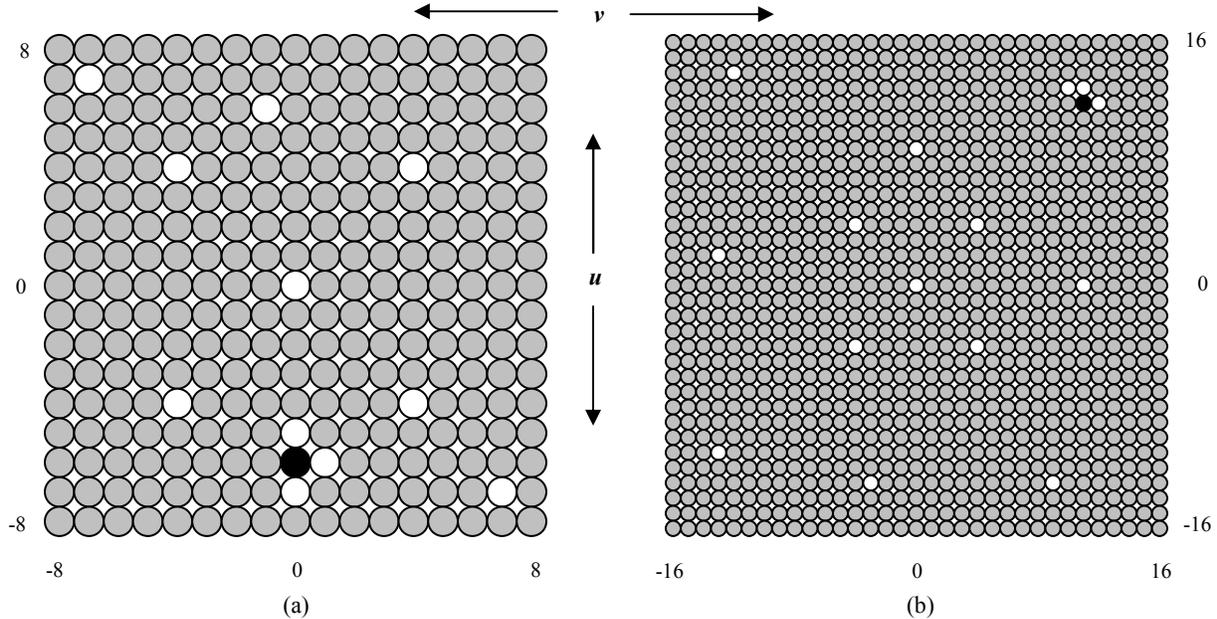

**Fig. 7.** Search-patterns generated by the ABC-BM algorithm. (a) Search window pattern ±8 with solution $\hat{u}^1_{best} = 0$ and $\hat{v}^1_{best} = -6$. (b) Search window pattern ±16 with solution $\hat{u}^2_{best} = 11$ and $\hat{v}^2_{best} = 12$

## 6. Experimental results

*6.1 ABC-BM results*

This section presents the results of comparing the proposed ABC-BM algorithm with other existing fast BM algorithms. Simulations have been performed over the luminance component of popular video sequences that are listed in Table 1. Such sequences consist of different degrees and types of motion including QCIF (176x144), CIF (352x288) and SIF (352x240) respectively. The first four sequences are *Container*, *Carphone*, *Foreman* and *Akiyo* in QCIF format. The next two sequences are *Stefan* in CIF format and *Football* in SIF format. Among such sequences, *Container* has gentle, smooth and low motion changes and consists mainly of stationary and quasi-stationary blocks. *Carphone*, *Foreman* and *Akiyo* have moderately complex motion getting a ''medium'' category regarding its motion content. Rigorous motion which is based on camera panning with translation and complex motion content can be found in sequences of *Stefan* and *Football*. Figure 8 shows a sample frame from each sequence.

Each picture frame is partitioned into macro-blocks with the sizes of 16x16 pixels for motion estimation, where the maximum displacement within the search space $W$ is ±8 pixels in both horizontal and vertical directions for the sequences *Container*, *Carphone*, *Foreman*, *Akiyo* and *Stefan*. The sequence *Football* has been simulated with a window size $W = \pm 16$ and requires a greater number of iterations (8 iterations) by the ABC-BM method.

In order to compare the performance of the ABC-BM approach, different search algorithms such as FSA, TSS [8], 4SS [11], NTSS [9], BBGD [15], DS [12], NE [16], ND [18], LWG [25], GFSS [26] and PSO-BM [27] have been all implemented in our simulations. For comparison purposes, all six video sequences in Fig. 8 have been all used. All simulations are performed on a Pentium IV 3.2 GHz PC with 1GB of memory.

In the comparison, two relevant performance indexes have been considered: the distortion performance and the search efficiency.

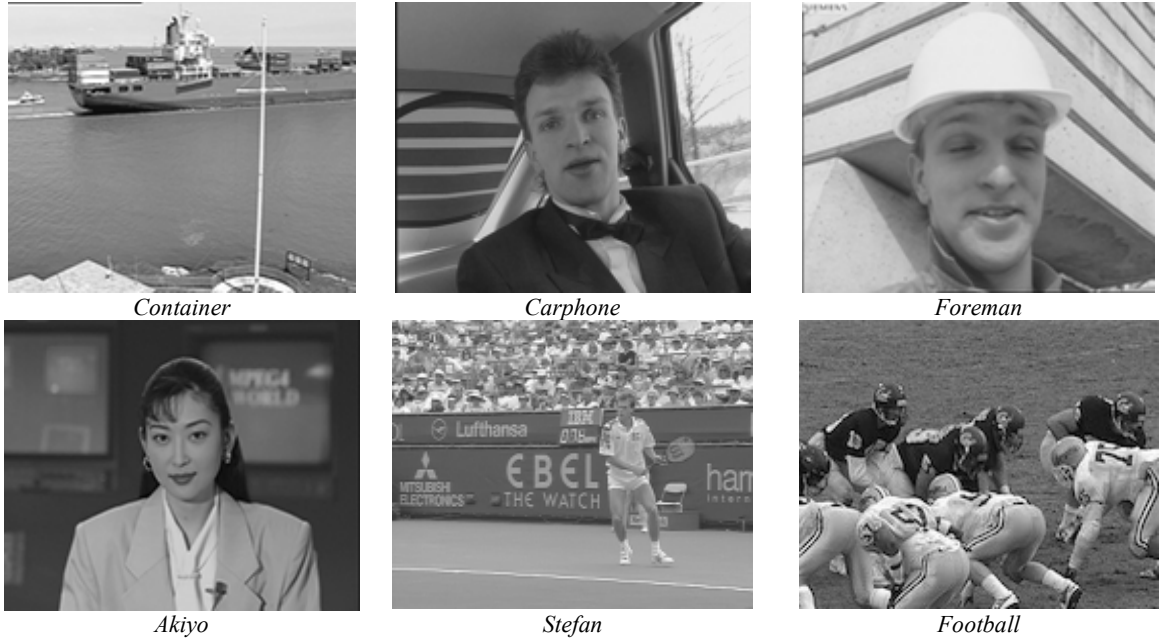

*Container*     *Carphone*     *Foreman*

*Akiyo*     *Stefan*     *Football*

**Fig. 8.** Test video sequences.

**Table 1.** Test sequences used in the comparison test.

| Sequence | Format | Total frames | Motion type |
|---|---|---|---|
| *Container* | QCIF(176x144) | 299 | Low |
| *Carphone* | QCIF(176x144) | 381 | Medium |
| *Foreman* | QCIF(352x288) | 398 | Medium |
| *Akiyo* | QCIF(352x288) | 211 | Medium |
| *Stefan* | CIF(352x288) | 89 | High |
| *Football* | SIF(352x240) | 300 | High |

Distortion performance

First, all algorithms are compared in terms of their distortion performance which is characterized by the Peak-Signal-to-Noise-Ratio (PSNR) value. Such value indicates the reconstruction quality when motion vectors, which are computed through a BM approach, are used. In PSNR, the signal comes from original data frames whereas the noise is the error introduced by the calculated motion vectors. The PSNR is thus defined as:

$$\text{PSNR} = 10 \cdot \log_{10}\left(\frac{255^2}{MSE}\right) \qquad (8)$$

where *MSE* is the mean square between the original frames and those compensated by the motion vectors. Additionally, as an alternative performance index, the PSNR degradation ratio ($D_{\text{PSNR}}$) is used in the comparison. This ratio expresses in percentage (%) the level of mismatch between the PSNR of a BM approach and the PSNR of the FSA which is considered as a reference. Thus the $D_{\text{PSNR}}$ is defined as:

$$D_{\text{PSNR}} = -\left(\frac{\text{PSNR}_{\text{FSA}} - \text{PSNR}_{\text{BM}}}{\text{PSNR}_{\text{FSA}}}\right) \cdot 100\% \tag{9}$$

**Table 2.** PSNR values and $D_{\text{PSNR}}$ comparison of the BM methods

| Algorithm | Container $W=\pm 8$ | | Carphone $W=\pm 8$ | | Foreman $W=\pm 8$ | | Akiyo $W=\pm 8$ | | Stefan $W=\pm 8$ | | Football $W=\pm 16$ | | Total Average |
|---|---|---|---|---|---|---|---|---|---|---|---|---|---|
| | PSNR | $D_{\text{PSNR}}$ | PSNR | $D_{\text{PSNR}}$ | PSNR | $D_{\text{PSNR}}$ | PSNR | $D_{\text{PSNR}}$ | PSNR | $D_{\text{PSNR}}$ | PSNR | $D_{\text{PSNR}}$ | ($D_{\text{PSNR}}$) |
| FSA | 43.18 | 0 | 31.51 | 0 | 31.69 | 0 | 29.07 | 0 | 25.95 | 0 | 23.07 | 0 | 0 |
| TSS | 43.10 | -0.20 | 30.27 | -3.92 | 29.37 | -7.32 | 26.21 | -9.84 | 21.14 | -18.52 | 20.03 | -13.17 | -8.82 |
| 4SS | 43.12 | -0.15 | 30.24 | -4.01 | 29.34 | -7.44 | 26.21 | -9.84 | 21.41 | -17.48 | 20.10 | -12.87 | -8.63 |
| NTSS | 43.12 | -0.15 | 30.35 | -3.67 | 30.56 | -3.57 | 27.12 | -6.71 | 22.52 | -13.20 | 20.21 | -12.39 | -6.61 |
| BBGD | 43.14 | -0.11 | 31.30 | -0.67 | 31.00 | -2.19 | 28.10 | -3.33 | 25.17 | -3.01 | 22.03 | -4.33 | -2.27 |
| DS | 43.13 | -0.13 | 31.26 | -0.79 | 31.19 | -1.59 | 28.00 | -3.70 | 24.98 | -3.73 | 22.35 | -3.12 | -2.17 |
| NE | 43.15 | -0.08 | 31.36 | -0.47 | 31.23 | -1.47 | 28.53 | -2.69 | 25.22 | -2.81 | 22.66 | -1.77 | -1.54 |
| ND | 43.15 | -0.08 | 31.35 | -0.50 | 31.20 | -1.54 | 28.32 | -2.56 | 25.21 | -2.86 | 22.60 | -2.03 | -1.59 |
| LWG | 43.16 | -0.06 | 31.40 | -0.36 | 31.31 | -1.21 | 28.71 | -1.22 | 25.41 | -2.09 | 22.90 | -0.73 | -0.95 |
| GFSS | 43.15 | -0.06 | 31.38 | -0.40 | 31.29 | -1.26 | 28.69 | -1.28 | 25.34 | -2.36 | 22.92 | -0.65 | -1.01 |
| PSO-BM | 43.15 | -0.07 | 31.39 | -0.38 | 31.27 | -1.34 | 28.65 | 1.42 | 25.39 | -2.15 | 22.88 | -0.82 | -1.03 |
| ABC-BM | 43.17 | **-0.02** | 31.50 | **-0.02** | 31.62 | **-0.22** | 29.02 | **-0.17** | 25.90 | **-0.19** | 23.02 | **-0.21** | **-0.17** |

Table 2 shows the comparison of PSNR values and PSNR degradation ratios ($D_{\text{PSNR}}$) among BM algorithms. The experiment considers all six image sequences presented in Fig. 8. It is evident at the slow-moving sequence *Container*, that the PSNR values (the $D_{\text{PSNR}}$ ratios) of all BM algorithms are similar. For the medium motion content sequences such as *Carphone*, *Foreman* and *Akiyo*, the approaches consistent of fixed patterns (TSS, 4SS and NTSS) exhibit the worst PSNR value (high $D_{\text{PSNR}}$ ratio) except for the DS algorithm. On the other hand, BM methods that use evolutionary algorithms (LWG, GFSS, PSO-BM and ABC-BM) present the lowest $D_{\text{PSNR}}$ ratio, only one step under the FSA approach which is considered as a reference. Finally, approaches based on the error-function minimization (BBGD and NE) and pixel-decimation (ND), show an acceptable performance. For the high motion sequence of *Stefan*, since the motion content of such sequences is complex and produces error surfaces with more than one minimum, the performance, in general, becomes worst for most of the algorithms in particular for those using fixed patterns. In the sequence *Football,* which has been simulated with a window size of ±16, the methods based on the evolutionary algorithms present the best PNSR values. Such performance is related to the fact that evolutionary methods adapt better to complex optimization problems with a bigger search area and a higher number of local minima. As a summary of the distortion performance, the last column of Table 2 presents the average PSNR degradation ratio ($D_{\text{PSNR}}$) from all sequences. According to such values, the proposed ABC-BM method is superior to any other approach. Due to its computation complexity, the FSA is considered just as a reference. Best entries are bold-cased in Table 2.

Search efficiency

The search efficiency is used in this section as a measurement of computational complexity. The search efficiency is calculated by counting the average number of search points (or the average number of SAD computations) for the MV estimation. In Table 3, the search efficiency is compared and the best results are bold-cased. Just above FSA, some evolutionary algorithms such as LWG, GFSS and PSO-BM hold the highest number of search points per block. In contrast, the proposed ABC-BM algorithm can be considered as a fast approach as it maintains a similar performance to DS. From data shown in Table 3, the average number of search locations that correspond to the ABC-BM method represents the number of SAD evaluations (the number of SAD estimations is not considered whatsoever). Additionally, the last two columns of Table 3 present the number of search locations that have been averaged (over the six sequences) and their performance rank. According to these values, the proposed ABC-BM method is ranked at the first place. The average number of search points visited by the ABC-BM algorithm ranges from 9.0 to 16.3, representing the 4% and the 7.4% respectively in comparison to the FSA method. Such results demonstrate that our approach can significantly reduce the number of search points. Hence, the ABC-BM algorithm proposed at this paper is comparable to the fastest algorithms and delivers a similar precision to the FSA approach.

**Table 3.** Averaged number of visited search points per block for all ten BM methods.

| Algorithm | Container $W=\pm 8$ | Carphone $W=\pm 8$ | Foreman $W=\pm 8$ | Akiyo $W=\pm 8$ | Stefan $W=\pm 8$ | Football $W=\pm 16$ | Total Average | Rank |
|---|---|---|---|---|---|---|---|---|
| FSA | 289 | 289 | 289 | 289 | 289 | 1089 | 422.3 | 12 |
| TSS | 25 | 25 | 25 | 25 | 25 | 25 | 25 | 8 |
| 4SS | 19 | 25.5 | 24.8 | 27.3 | 25.3 | 25.6 | 24.58 | 7 |
| NTSS | 17.2 | 21.8 | 22.1 | 23.5 | 25.4 | 26.5 | 22.75 | 6 |
| BBGD | 9.1 | 14.5 | 14.5 | 13.2 | 17.2 | 22.3 | 15.13 | 3 |
| DS | 7.5 | 12.5 | 13.4 | 11.8 | **15.2** | 17.8 | 13.15 | 2 |
| NE | 11.7 | 13.8 | 14.2 | 14.5 | 19.2 | 24.2 | 16.36 | 5 |
| ND | 10.8 | 13.4 | 13.8 | 14.1 | 18.4 | 25.1 | 16.01 | 4 |
| LWG | 75 | 75 | 75 | 75 | 75 | 75 | 75 | 11 |
| GFSS | 60 | 60 | 60 | 60 | 60 | 60 | 60 | 10 |
| PSO-BM | 32.5 | 48.5 | 48.1 | 48.5 | 52.2 | 52.2 | 47 | 9 |
| ABC-BM | **9.0** | **11.2** | **10.2** | **12.5** | 16.1 | **16.3** | **12.14** | **1** |

*6.2 Results on H.264*

In order to evaluate the performance of the proposed algorithm, a set of experiments has been developed for JM-12.2 [51] of H.264/AVC reference software. Considering such encoder profile, the test conditions have been set as follows: For each test sequence only the first frame has been coded as "I frame" while remaining frames are coded as "P frames". Only one reference frame has been used. The sum of absolute differences (SAD) distortion function is employed as the block distortion measure. The simulation platform in our experiments is a PC with Intel Pentium IV 2.66 GHz CPU.

Test sequences used in the experiments are *Container*, *Akiyo* and *Football*. Such sequences exhibit a variety of motion that is generally encountered in real video. A search window of ±8 is selected for sequences *Container* and *Akiyo* while the *Football* sequence is processed through a search window of ±16. Image formats used by the sequences are QCIF, CIF and SIF, testing at 30 fps over 200 different frames.

The group of experiments has been performed over test sequences at four different quantization parameters (QP=28,32,36,40) in order to test the algorithm at different bit rates. In the simulations, we have compared DS [12], EPZS [20] and the proposed ABC-BM algorithm against the FSA, in terms of coding efficiency and computational complexity. The FSA is used as the basis of the image quality.

For the evaluation of coding efficiency, the Bjontegaard Delta PSNR (BDPSNR) and the Bjontegaard Delta Bit-Rate (BDBR) are used as performance indexes. Such indexes objectively express the average differences of PSNR and Bit-Rate when two methods are compared [52]. In order to calculate the computational complexity, the speed up ratio (SUR) is employed as computational efficiency index. SUR specifies the averaged speed up ratio between a given algorithm's motion estimation time and the time demanded by FSA.

Table 4 shows the comparative results of DS, EPZS and the proposed ABC-BM algorithm against the FSA. The sign (-) in BDPSNR and the sign (+) for BDBR indicate a loss in the coding performance. It is observed from experimental results that the proposed ABC-BM algorithm is effective for reducing the computational complexity of the motion estimation. The speed up ratio (SUR) is about 60 times faster than FSA and it is about 3 times faster than EPZS, preserving a similar computation load in comparison to DS. Likewise, the coding performance of the ABC-BM algorithm is efficient because the loss in terms of BDBR and BDPSNR is low with only 0.63% and 0.07 dB, respectively. Such coding performance is similar to the one produced by the EPZS method whereas it is better than the obtained by the DS algorithm.

Table 4. Performance comparison of DS, EPZS and ABC-BM for three different sequences in JM-12.2.

| Sequence | DS | | | EPZS | | | ABC-BM | | |
|---|---|---|---|---|---|---|---|---|---|
| | BDBR (%) | BDPSNR (dB) | SUR | BDBR (%) | BDPSNR (dB) | SUR | BDBR (%) | BDPSNR (dB) | SUR |
| *Container* $W = \pm 8$. | +4.22 | -0.48 | 50.33 | +0.33 | -0.02 | 28.12 | +0.35 | -0.02 | 65.87 |
| *Akiyo* $W = \pm 8$. | +6.15 | -0.54 | 44.87 | +0.64 | -0.07 | 24.79 | +0.67 | -0.08 | 58.14 |
| *Football* $W = \pm 16$ | +7.11 | -0.56 | 48.78 | +0.81 | -0.11 | 25.12 | +0.88 | -0.12 | 60.47 |
| Average | **+5.82** | **-0.52** | **48.00** | **+0.59** | **-0.06** | **26.01** | **+0.63** | **-0.07** | **61.49** |

*Experiments with high definition sequences*

This section presents yet more experiments in the JM-12.2 encoder profile in order to evaluate the performance of the proposed algorithm over high definition sequences. Three testing video sequences are taken into consideration in the experiments. As it is shown in Table 5, such testing sequences include video data of various resolutions and different motion activities. Figure 9 presents a sample frame from each sequence.

Table 5. High definition test sequences used in the comparison.

| Sequence | Resolution | Total frames | Motion type |
|---|---|---|---|
| *Blue-sky* | 1024x786 | 200 | Low |
| *Pedestrian-Area* | 1024x786 | 200 | Medium |
| *Rush-field* | 640x480 | 200 | High |

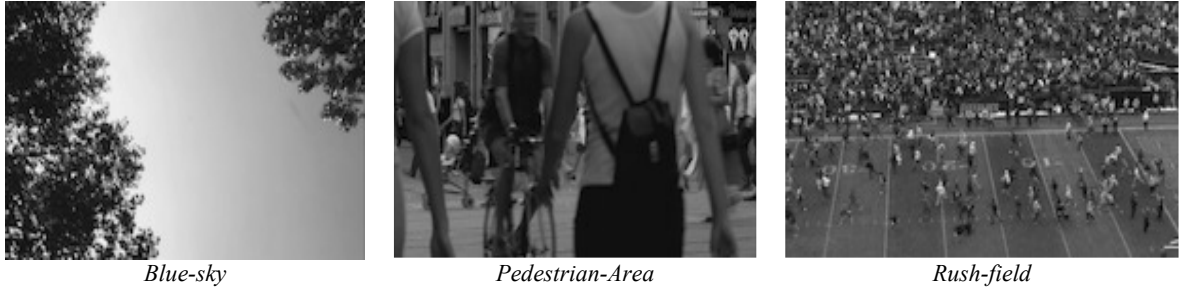

*Blue-sky*  *Pedestrian-Area*  *Rush-field*

**Fig. 9.** High definition test video sequences.

For encoding purposes, the JM-12.2 encoder profile has been set by using the first frame which has been coded as "I frame" and the remaining frames which have been coded as "P frames". Similarly to previous experiments, only one reference frame has been employed and the sum of absolute differences (SAD) distortion function is used as the block distortion measure. All sequences are tested at 30 fps considering 200 different frames. The search range for the sequences is ±32 whereas four QP (28,32,36,40) values are used for calculating the Bjontegaard Delta PSNR (BDPSNR) and the Bjontegaard Delta Bit-Rate (BDBR).

Table 6 shows the performance comparison of the proposed ABC-BM algorithm with DS [12] and EPZS [20] while using the FSA method as the basis of the image quality. As it is shown in Table 6, the DS algorithm possesses the worst coding efficiency with 8.1% and 0.7 dB respectively. However, it also presents a competitive SUR of 56.65. The EPZS method has the best coding performance (0.51% and 0.04 dB), holding the worst speed up ratio at 21.13. On the other hand, the proposed ABC-BM algorithm maintains the best trade off between coding efficiency (0.53% and 0.05 dB) and computational complexity (59.5).

**Table 4.** Performance comparison of DS, EPZS and ABC-BM for three different high definition sequences in JM-12.2.

| Sequence | DS | | | EPZS | | | ABC-BM | | |
|---|---|---|---|---|---|---|---|---|---|
| | BDBR (%) | BDPSNR (dB) | SUR | BDBR (%) | BDPSNR (dB) | SUR | BDBR (%) | BDPSNR (dB) | SUR |
| *Blue-sky* | +7.05 | -0.61 | 60.91 | +0.21 | -0.01 | 26.11 | +0.22 | -0.02 | 64.22 |
| *Pedestrian-Area.* | +8.24 | -0.72 | 57.14 | +0.72 | -0.04 | 24.12 | +0.73 | -0.05 | 60.04 |
| *Rush-field* | +9.03 | -0.79 | 51.92 | +0.62 | -0.08 | 19.17 | +0.66 | -0.10 | 54.21 |
| Average | **+8.10** | **-0.70** | **56.65** | **+0.51** | **-0.04** | **23.13** | **+0.53** | **-0.05** | **59.50** |

## 7. Conclusions

In this paper, a new algorithm based on Artificial Bee Colony (ABC) is proposed to reduce the number of search locations in the BM process. The algorithm uses a simple fitness calculation approach which is based on the Nearest Neighbor Interpolation (NNI) algorithm. The method is able to save computational time by identifying which fitness value can be just estimated or must be calculated instead. As a result, the approach can substantially reduce the number of function evaluations, yet preserving the good search capabilities of ABC.

Since the proposed algorithm does not consider any fixed search pattern or any other movement assumption, a high probability for finding the true minimum (accurate motion vector) is expected regardless of the movement complexity contained in the sequence. Therefore, the chance of being trapped into a local minimum is reduced in comparison to other BM algorithms.

The performance of ABC-BM has been compared to other existing BM algorithms by considering different sequences which present a great variety of formats and movement types. Experimental results demonstrate the high performance of the proposed method in terms of coding efficiency and computational complexity.

Although the experimental results indicate that the ABC-BM method can yield better results on complicated sequences, it should be noticed that the aim of our paper is not intended to beat all the BM methods which have been proposed earlier, but to show that the fitness approximation can effectively serve as an attractive alternative to evolutionary algorithms for solving complex optimization problems, yet demanding fewer function evaluations.